\documentclass[runningheads]{llncs}
\usepackage[T1]{fontenc}
\usepackage{graphicx}
\usepackage{hyperref}
\usepackage{color}

\usepackage{amsmath}
\usepackage{multirow}
\usepackage{multicol}
\usepackage{colortbl}
\usepackage[table]{xcolor}
\usepackage{graphicx}
\usepackage{booktabs}
\usepackage[ruled,vlined]{algorithm2e} 
\usepackage[misc]{ifsym}

\begin{document}
\title{Diffusion-based Hierarchical Negative Sampling for Multimodal Knowledge Graph Completion}
\titlerunning{Diffusion-based Hierarchical Negative Sampling for MMKGC}
%
\author{Guanglin Niu\inst{1(\textrm{\Letter})} \and
Xiaowei Zhang\inst{2}}
\authorrunning{G. Niu et al.}
%
\institute{School of Artificial Intelligence, Beihang University, Beijing, China
\email{beihangngl@buaa.edu.cn} \and
College of Computer Science and Technology, Qingdao University, Qingdao, China\\
\email{xiaowei19870119@sina.com}}
\maketitle              
\begin{abstract}
Multimodal Knowledge Graph Completion (MMKGC) aims to address the critical issue of missing knowledge in multimodal knowledge graphs (MMKGs) for their better applications. However, both the previous MMGKC and negative sampling (NS) approaches ignore the employment of multimodal information to generate diverse and high-quality negative triples from various semantic levels and hardness levels, thereby limiting the effectiveness of training MMKGC models. Thus, we propose a novel Diffusion-based Hierarchical Negative Sampling (DHNS) scheme tailored for MMKGC tasks, which tackles the challenge of generating high-quality negative triples by leveraging a Diffusion-based Hierarchical Embedding Generation (DiffHEG) that progressively conditions on entities and relations as well as multimodal semantics. Furthermore, we develop a Negative Triple-Adaptive Training (NTAT) strategy that dynamically adjusts training margins associated with the hardness level of the synthesized negative triples, facilitating a more robust and effective learning procedure to distinguish between positive and negative triples. Extensive experiments on three MMKGC benchmark datasets demonstrate that our framework outperforms several state-of-the-art MMKGC models and negative sampling techniques, illustrating the effectiveness of our DHNS for training MMKGC models. The source codes and datasets of this paper are available at \url{https://github.com/ngl567/DHNS}.

\end{abstract}
\section{Introduction}



Multimodal knowledge graphs (MMKGs) have become a powerful paradigm for representing symbolic knowledge, integrating diverse modalities such as text, images, and audio~\cite{MMKGSurvey}. These graphs are extensively applied across various domains such as multimodal question answering systems~\cite{MMKGQA}, where they enrich contextual relevance by representing multimodal information.

Knowledge graph completion (KGC) is an essential task in the context of MMKGs, as real-world MMKGs are frequently incomplete due to constraints in data collection and curation~\cite{MMKG}. The objective of MMKGC is to infer missing knowledge and then enhance the MMKG's completeness and utility. In the training of MMKGC models, negative sampling (NS) is a critical component, given the scarcity of negative triples in any MMKG~\cite{MMRNS}. NS generates negative triples that contrast with the positive triples in MMKGs, allowing the model to learn semantic boundaries and associations among entities and relations.


However, existing negative sampling (NS) strategies, such as random sampling and adversarial-based methods, face three key challenges. First, \textbf{Current NS techniques~\cite{NSsurvey} mainly rely on topological features while neglecting semantics from diverse modalities} especially in MMKGs, leading to simple or invalid negative triples. Second, although some approaches based on generative adversarial networks (GANs)~\cite{kbgan} or self-adversarial strategies~\cite{RotatE} can assess the quality of sampled negative triples, \textbf{their assessment depends on pre-sampled triples and the performance of knowledge graph completion (KGC) models, rather than directly generating high-quality negative triples}. Third, current KGC models employ a fixed margin for training~\cite{KGCSurvey}, \textbf{making it challenging for a one-margin-fits-all training scheme to be effective across different hardness levels of negative triples}.

To address these challenges, we propose a novel \underline{\textbf{D}}iffusion-based \underline{\textbf{H}}ierarchical \underline{\textbf{N}}egative \underline{\textbf{S}}ampling (DHNS) paradigm for MMKGC, which is motivated by the recent success of diffusion models in various generative tasks. By leveraging the powerful denoising diffusion probabilistic model (DDPM)~\cite{DDPM}, we develop a \textbf{Diffusion-based Hierarchical Embedding Generation (DiffHEG)} module, which could directly generate diverse entity embeddings for composing negative triples, instead of the traditional NS paradigm of sampling entities. Specifically, the synthesized negative triples are obtained concerning hierarchical semantics from both multiple modality-specific embeddings and various hardness levels via conditional denoising at different time steps. Furthermore, these high-quality and diverse negative triples could be fed into any KGC model, enhancing its ability to distinguish between positive and negative triples. Specifically, we develop a \textbf{Negative Triple-Adaptive Training (NTAT)} mechanism with Hardness-Adaptive Loss (HAL) to enhance the KGC model's learning capability for different hardness levels of negative triples. An architecture of our DHNS framework is illustrated in Fig.~\ref{fig:framework}. Our contributions can be summarized as:

\begin{figure}
    \centering
    \includegraphics[width=1\linewidth]{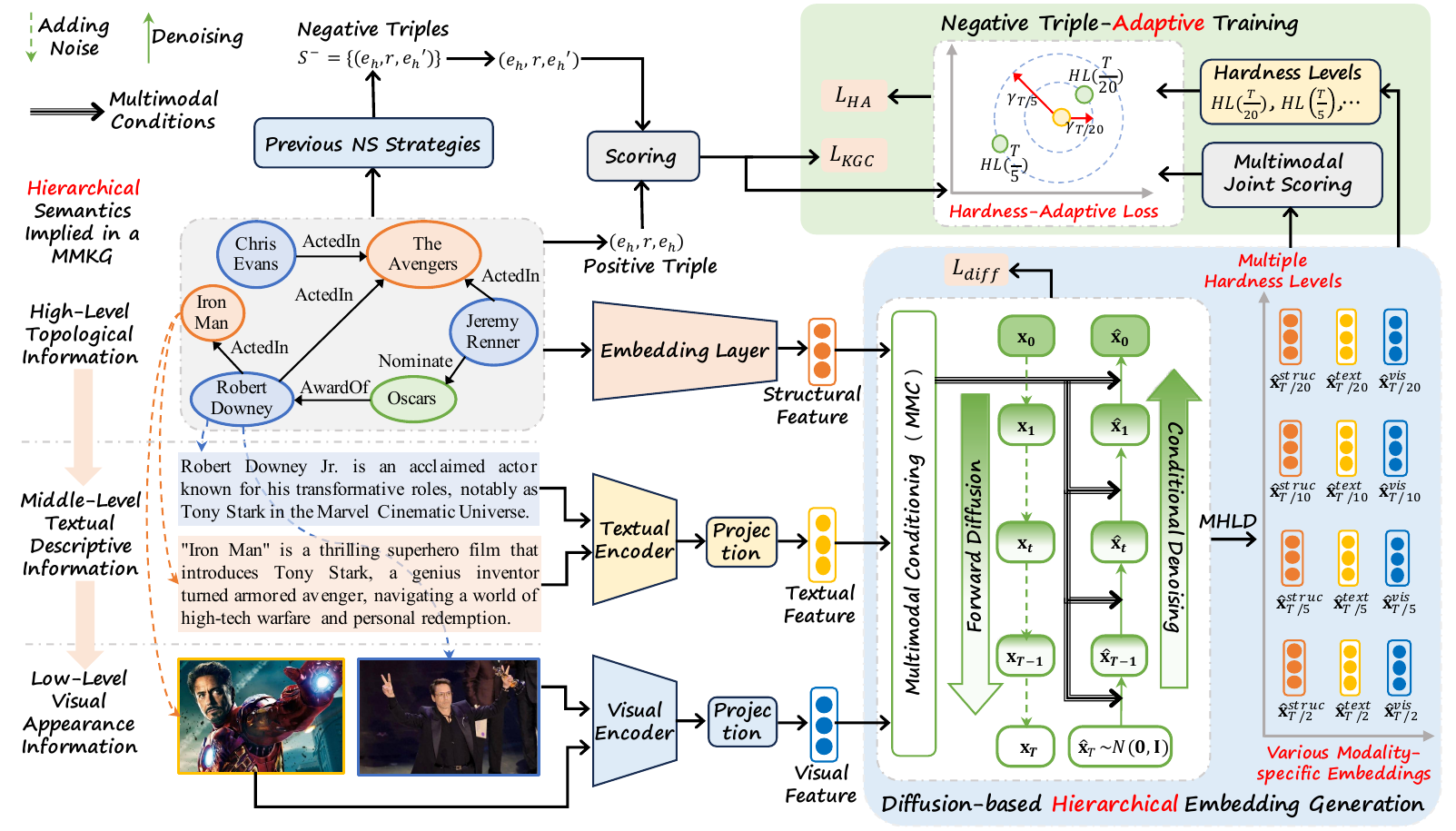}
    \caption{The whole framework of our DHNS. MHLD means multiple hardness-level denoising. $\textbf{x}_{0:T}$ and $\hat{\textbf{x}}_{0:T}$ are the noised and the denoised embeddings in the range of time steps $[0, T]$ corresponding to an entity. $\textbf{x}_{T/20}^{struc}$, $\textbf{x}_{T/20}^{text}$ and $\textbf{x}_{T/20}^{vis}$ are three modality-specific (structural/textual/visual) denoised embeddings at the time step $T/20$. $\gamma_{T/20}$ denotes the margin adaptive to the negative triples with the hardness level $HL(\frac{T}{20})$.}
    \label{fig:framework}
\end{figure}

\begin{itemize}
    \item As we can be concerned, it is the first effort to leverage the diffusion model's capabilities within the context of MMKGC for negative sampling. Our NS module DiffHEG captures the diverse semantics of different modalities to generate hierarchical and high-quality negative triples while directly controlling the hardness levels with diffusion time steps.
    \item Based on the generated negative triples, we develop a hardness-adaptive training objective in which the pivotal parameter margin is adaptive to hardness levels of negative triples. It facilitates the comprehensive training of an MMKGC model for a diverse range of negative triples.
    \item Extensive experiments are conducted on three MMKGC benchmark datasets to compare the performance of our DHNS model against some state-of-the-art MMKGC models and negative sampling strategies, demonstrating the robustness and effectiveness of our DHNS framework for MMKGC tasks.
\end{itemize}

\section{Related Work}

\subsection{Multimodal Knowledge Graph Completion}

Existing MMKGC approaches extend traditional knowledge graph embedding (KGE) techniques to handle multi-modal information~\cite{Xie:IKRL}. KGE models represent entities and relations in continuous numerical spaces. Early models like TransE~\cite{Bordes:TransE} propose a translational distance-based score function, where relations are represented as translation operations between entities. DistMult~\cite{Distmult} and ComplEx~\cite{Trouillon:ComplEx} use bilinear models to capture both symmetric and antisymmetric relations. RotatE~\cite{RotatE} and QuatE~\cite{QuatE} introduce rotational and quaternion-based embeddings, respectively, to model complex relational patterns.

In particular, MMKGC models typically involve designing additional embeddings to represent multi-modal information, such as textual descriptions and images of entities, and incorporating them into the score function to evaluate the plausibility of each triple. For instance, IKRL~\cite{Xie:IKRL} extracts visual features using a pre-trained visual encoder and combines them with structural embeddings from TransE to assess the plausibility of triples. TransAE~\cite{TransAE} and TBKGC~\cite{TBKGC} extend IKRL by integrating both visual and textual information. RSME~\cite{RSME} employs a gate mechanism to ensure that the most relevant multi-modal features are incorporated into entity embeddings. AdaMF~\cite{AdaMF} uses a generator to produce adversarial samples and a discriminator to measure their plausibility in an adversarial training framework. LAFA~\cite{LAFA} considers the relationships between entities and different modalities, focusing on link-aware aggregation of multi-modal information. VISTA~\cite{VISTA} designs three transformer-based encoders to incorporate visual and textual embeddings for predicting missing triples. However, these MMKGC models primarily focus on entity and relation representation learning, neglecting the importance of generating high-quality negative triples from rich multi-modal information to better guide the training process.


\subsection{Negative Sampling in Knowledge Graph Embedding}

Negative sampling (NS)\cite{NSsurvey} is a widely used technique in KGE that generates negative triples not present in knowledge graphs (KGs), enhancing model training by contrasting them with positive triples. Traditional strategies, such as random entity replacement, are simple but often produce false negatives or low-quality triples, leading to ambiguous training signals~\cite{kbgan}. Bernoulli sampling~\cite{Wang:TransH} employs a Bernoulli distribution to replace entities to generate higher-quality negative triples. KBGAN~\cite{kbgan} and IGAN~\cite{IGAN} use Generative Adversarial Networks (GANs) to select harder negatives that are difficult for KGE models to distinguish from positives. NSCaching~\cite{NSCaching} utilizes additional memory to store and efficiently sample high-quality negative triples during training. SANS~\cite{SANS} leverages graph structure information for sampling high-quality negatives. However, these NS strategies are primarily designed for unimodal KGC models and do not leverage the multi-level semantics in multimodal information, which is crucial for generating diverse negative triples. MMRNS~\cite{MMRNS} introduces a relation-enhanced NS mechanism using knowledge-guided cross-modal attention to generate more challenging negatives from multimodal data. MANS~\cite{MANS} emphasizes modality-aware NS to align structural and multimodal information, enhancing negative triple quality. Despite these advances, these multimodal NS approaches remain within the sampling paradigm, lacking control over negative triple generation.


\subsection{Diffusion Models}

In recent years, diffusion models, particularly the Denoising Diffusion Probabilistic Model (DDPM)~\cite{DDPM}, have become pivotal in AI-generated content by progressively adding Gaussian noise to input data across time steps, forming a Markov chain based on a noise schedule. The reverse process involves training a neural network to predict and remove noise, thereby reconstructing the original data. In graph-structured data, DMNS\cite{DMNS} employs DDPM to generate negative nodes for link prediction, considering the query node's context, while KGDM~\cite{KGDM} applies DDPM to estimate the probabilistic distribution of target entities for KGC. However, the application of diffusion models for NS in KGC, particularly in MMKGC, remains limited. Our work addresses this gap by leveraging diffusion models for negative sampling in MMKGC, adapting the model to MMKGs to generate diverse negative triples of varying hardness, capturing hierarchical multi-modal semantics and enhancing MMKGC model training.

\section{Methodology}

\subsection{Preliminaries and Problem Definition}

A multimodal knowledge graph (MMKG) extends the traditional knowledge graph (KG) and is defined as a tuple $\mathcal{G} = (\mathcal{E}, \mathcal{R}, \mathcal{T}, \mathcal{M})$. Here, $\mathcal{E}$ and $\mathcal{R}$ denote the sets of entities and relations, and $\mathcal{T}$ is the set of triples in the form $(e_h, r, e_t)$, where $e_h, e_t \in \mathcal{E}$ and $r \in \mathcal{R}$. The component $\mathcal{M}$ indicates multimodal information, including images and textual descriptions associated with entities.


The objective of MMKGC is to predict missing triples in $\mathcal{G}$. Given an incomplete triple query, such as $(e_h, r, ?)$ or $(?, r, e_t)$, where $e_h$ and $e_t$ are known entities and $r$ is a relation, the task is to evaluate the plausibility of candidate triples $(e_h, r, e_t)$ or $(e, r, e_t)$ by scoring them based on their learned embeddings in $\mathcal{G}$. During training, the primary goal is to distinguish between positive triples $(e_h, r, e_t)$ and their corresponding negative triples $(e_h', r, e_t)$ or $(e_h, r, e_t')$ where $e_h'$ and $e_t'$ are the corrupted entities sampled from $\mathcal{E}$. To enhance the model's discriminative capability, negative sampling aims to generate hard negative triples that are semantically similar to their positive counterparts. In this paper, we represent the original embedding of an entity or relation as $\textbf{x}$.

\subsection{Diffusion-based Hierarchical Embedding Generation}

To supplement the negative triples obtained by sampling explicit entities, we propose a novel Diffusion-based Hierarchical Embedding Generation (DiffHEG) module. This DHNS module enables the generation of hierarchical entity embeddings to compose high-quality negative triples by conditioning on entities and relations as well as multimodal information. Besides, these negative triples vary multiple levels of hardness, which are regulated by diffusion time steps. 

\subsubsection{Forward Diffusion and Reverse Denoising Procedures.}

Following the basic architecture of DDPM, in the forward diffusion process, the input entity embedding $\textbf{x}_0$ gradually has noise added to it, resulting in a sequence of embeddings that converge to pure Gaussian noise $\textbf{x}_T$, in which $T$ indicates the total time steps of the diffusion process. Specifically, the forward diffusion process can be represented by a Markov chain as follows:
\begin{equation}
    q(\textbf{x}_t|\textbf{x}_{t - 1}) = \mathcal{N}(\textbf{x}_t; \sqrt{1 - \alpha_t} \textbf{x}_{t - 1}, \alpha_t \textbf{I})
\end{equation}
where $\alpha_t$ represents the variance that could be constants or learnable by some scheduling mechanisms at arbitrary time step $t$. The complete process of adding noise can be expressed as $q(\textbf{x}_{1:T}|\textbf{x}_0) = \prod_{t = 1}^{T}q(\textbf{x}_t|\textbf{x}_{t - 1})$. Thus, the closed form of noisy entity embedding $\textbf{x}_t$ is calculated by
\begin{equation}
    \textbf{x}_t = \sqrt{\bar{\beta}_t} \textbf{x}_0 + \sqrt{1-\bar{\beta}_t} \epsilon_t \label{eq3}
\end{equation}
where $\beta_t = 1-\alpha_t$ and $\bar{\beta}_t=prod_{i = 1}^{t}\beta_i$. $\epsilon_t \sim \mathcal{N}(\textbf{0}, \textbf{I})$ is the noise.

Furthermore, the reverse diffusion process iteratively denoises the noisy entity embedding $\textbf{x}_t$ to obtain the synthetic entity embedding. Given a noisy entity embedding $\textbf{x}_t$ at time step $t$, the reverse process is conditioned on the embeddings of the observed entity $\textbf{x}_h$ and the relation $\textbf{x}_r$, yielding:
\begin{equation}
    p_\theta(\textbf{x}_{t - 1}|\textbf{x}_t) = \mathcal{N}(\textbf{x}_{t - 1}; \mu_{\theta}(\textbf{x}_t, PE(t), C(\textbf{x}_e, \textbf{x}_r)), \alpha_t \textbf{I}) \label{eq4}
\end{equation}
where $\theta$ denotes the parametrization of the diffusion model. $PE(t)$ denotes the positional embedding of the time step $t$ and will be declared in the following, and $C(\textbf{x}_e, \textbf{x}_r)$ indicates the condition derived from an observed entity $e$ and the associated relation $r$ in the triple for guiding the generation of another entity embedding to constitute a negative triple. Particularly, $\mu_{\theta}(\textbf{x}_t,$ $PE(t), C(\textbf{x}_e, \textbf{x}_r))$ indicates the mean of the Gaussian distribution and is parameterized as:
\begin{equation}
    \mu_{\theta}(\textbf{x}_t, PE(t), C(\textbf{x}_e, \textbf{x}_r)) = \frac{1}{\sqrt{\beta_t}} \textbf{x}_t - \frac{\beta_t}{\sqrt{\beta_t} \sqrt{1-\bar{\beta}_t}}\epsilon_{\theta}(\hat{\textbf{x}}_t, PE(t), C(\textbf{x}_e, \textbf{x}_r)) \label{eq5}
\end{equation}
in which $\epsilon_{\theta}(\hat{\textbf{x}}_t, PE(t), C(\textbf{x}_e, \textbf{x}_r)$ is the estimated noise obtained through the following conditional denoising operation. From Eq.~\ref{eq4} and Eq.\ref{eq5}, the generated entity embedding via eliminating predicted noises at the beginning and each intermediate time step in the reverse diffusion process could be rewritten as:
\begin{align}
    \hat{\textbf{x}}_{T} \sim \mathcal{N}(\textbf{0}, \textbf{I}),
    \hat{\textbf{x}}_{0} = \frac{1}{\sqrt{\beta_t}} \hat{\textbf{x}}_1 - \frac{\beta_t}{\sqrt{\beta_t} \sqrt{1-\bar{\beta}_t}}\epsilon_{\theta}(\hat{\textbf{x}}_1, PE(1), C(\textbf{x}_e, \textbf{x}_r)) \label{eq7} \\
    \hat{\textbf{x}}_{t-1} = \frac{1}{\sqrt{\beta_t}} \hat{\textbf{x}}_t - \frac{\beta_t}{\sqrt{\beta_t} \sqrt{1-\bar{\beta}_t}}\epsilon_{\theta}(\hat{\textbf{x}}_t, PE(t), C(\textbf{x}_e, \textbf{x}_r)) + \alpha_t \epsilon_t \label{eq8}
\end{align}
where $\hat{\textbf{x}}_{T}$ implies the pure noise at the last step $T$, $\hat{\textbf{x}}_{0}$ and $\hat{\textbf{x}}_{t-1}$ denote the generated entity embedding at the beginning step and the intermediate time step in the range of $[1, T-1]$. Following the idea of DDPM that promotes the denoise process with more diversity, the item $\alpha_t \epsilon_t$ is utilized to add a random noise for obtaining each intermediate denoise result $\hat{\textbf{x}}_{t-1}$.

\subsubsection{Multimodal Conditioning and Multiple Hardness-Level Denoising.}

Conditional denoising is composed of two sub-modules: Multimodal Conditioning (MMC) and Multiple Hardness-Level Denoising (MHLD). The MMC module computes a conditional embedding \( C(\mathbf{x}_e, \mathbf{x}_r) \) by integrating entity and relation embeddings across multiple semantic levels from structural, visual, and textual features. This integration captures rich contexts of entities in the MMKG, facilitating the denoising process. Given the various strategies for modeling triples in KGs, we use several condition calculation mechanisms, detailed as follows: 

Hardmard multiplication, which is available for rotation-based KGE models such as RotatE~\cite{RotatE} and QuatE~\cite{QuatE}, is formulated as:
\begin{equation}
    C(\textbf{x}_e, \textbf{x}_r) = \textbf{x}_e \circ \textbf{x}_r \label{eq9}
\end{equation}
where $\circ$ means Hardmard multiplication.

Bilinear multiplication, which is suitable for bilinear interaction-based KGE models such as RESCAL~\cite{RESCAL} and DistMult~\cite{Distmult}, formulated as:
\begin{equation}
    C(\textbf{x}_e, \textbf{x}_r) = \textbf{x}_e \times \textbf{x}_r \label{eq10}
\end{equation}
where $\times$ indicates element-wise multiplication.

Addition, which is available for translation-based KGE models such as TransE~\cite{Bordes:TransE} and TransH~\cite{Wang:TransH}, formulated as:
\begin{equation}
    C(\textbf{x}_e, \textbf{x}_r) = \textbf{x}_e + \textbf{x}_r \label{eq11}
\end{equation}

The MHLD module applies a denoising transformation on the noisy entity embedding $\textbf{x}_t$, informed by the conditional embedding $C(\textbf{x}_e, \textbf{x}_r)$ and the time embedding $PE(t)$. In particular, $PE(t)$ is modulated by a sinusoidal positional embedding layer that is frequently used in Transformer-based models to guarantee the temporal constraints among time steps and effectively guide the denoising at each time step $t$. For instance, $[PE(t)]_{2i}=sin(t/1000^{\frac{2i}{d}})$ and $[PE(t)]_{2i+1}=cos(t/1000^{\frac{2i}{d}})$ where $d$ is the dimension of $PE(t)$. Given the inputs of $\textbf{x}_t$, $PE(t)$, and $C(\textbf{x}_e, \textbf{x}_r)$, the denoising function can be represented as:
\begin{equation}
    \epsilon_{\theta}(\textbf{x}_t, PE(t), C(\textbf{x}_e, \textbf{x}_r)) = LayerNorm(MLP(\textbf{x}_t, PE(t), C(\textbf{x}_e, \textbf{x}_r))) \label{eq13}
\end{equation}
where simple multi-layer perceptron (MLP) and layer normalization layer (LayerNorm) are leveraged to predict the noise with the learnable parameters $\theta$.


The MMC module integrates triples, images, and texts to generate diverse, semantically rich negative triples. Specifically, by corrupting the tail entity of a positive triple $(e_h, r, e_t)$, we input the head entity's structural feature $\textbf{x}_{eh}^{struc}$, visual feature $\textbf{x}_{eh}^{vis}$, and textual feature $\textbf{x}_{eh}^{text}$, alongside the relation embedding $\textbf{x}_r$, into MHLD module to compute multimodal conditions. These conditions guide the reverse process outlined in Eqs.~\ref{eq7}-\ref{eq8} and \ref{eq13}, yielding various modality-specific embeddings $\hat{\textbf{x}}_{et,t}^{struc}$, $\hat{\textbf{x}}_{et,t}^{vis}$, and $\hat{\textbf{x}}_{et,t}^{text}$ for constructing the generated negative triples. The DHNS model generates multimodal implicit entity embeddings for the corrupted tail entity by minimizing a denoising diffusion loss:
\begin{align}
    \mathcal{L}_{diff} = &\|\epsilon_{\theta}(\hat{\textbf{x}}_{et,t}^{struc}, PE(t), C(\textbf{x}_{eh}^{struc}, \textbf{x}_r)) - \epsilon_t\|^2 + \|\epsilon_{\theta}(\hat{\textbf{x}}_{et,t}^{vis}, PE(t), \\ \nonumber
    & C(\textbf{x}_{eh}^{vis}, \textbf{x}_r))- \epsilon_t\|^2 + \|\epsilon_{\theta}(\hat{\textbf{x}}_{et,t}^{text}, PE(t), C(\textbf{x}_{eh}^{text}, \textbf{x}_r)) - \epsilon_t\|^2
\end{align}
where the denoising diffusion loss is formulated as the mean squared error (MSE) between the predicted noises from structural, textual, and visual levels and the actual noise added during the forward diffusion process. This loss is optimized using the Adam optimizer with separate learning rates to ensure stability.

The entity embeddings are generated by starting from complete noise and iteratively removing the predicted noise at each time step, a process inherent to diffusion models. To control the hardness of the negative triples, we modulate the diffusion time steps. Smaller time steps yield negatives closer to positives, representing higher hardness, while larger time steps produce easily distinguishable negatives. We formalize the hardness level of the generated entity embedding $\hat{\textbf{x}}_t$ as inversely proportional to the time step $t$, such as $HL(\hat{\textbf{x}}_t) \propto \frac{1}{t}$.

To ensure diversity, we sample time steps at specific intervals, such as \( t = T/20, T/10, T/5, T/2 \), to obtain a set of generated entity embeddings \( G^{-} = \{neg_t | t = T/20, T/10, T/5, T/2\} \), where \( neg_t = (\hat{\textbf{x}}_t^{struc}, \hat{\textbf{x}}_t^{vis}, \hat{\textbf{x}}_t^{text}) \) represents the modality-specific embeddings at time step \( t \). To balance quality, embeddings closer to the halfway point of the diffusion process are assigned higher weights.

The DiffHEG mechanism provides hierarchical control over negative triple generation, considering both hardness levels and modalities, offering a more nuanced approach than random replacements. By combining diverse negative triples, we enhance the training signal for the KGE model, potentially improving performance and generalization.

\subsection{Negative Triple-Adaptive Training}

Considering the varying hardness levels of negative affect the KGE model's ability to distinguish between positive and negative triples, we propose a well-designed Negative Triple-Adaptive Training (NTAT) mechanism for training KGE models based on the generated negative triples. Specifically, HTAT consists of two modules: multimodal joint scoring and Hardness-Adaptive Loss (HAL). The multimodal joint scoring computes a joint score for generated negative triples by integrating entity embeddings from structural, visual, and textual modalities. Take the generated negative triple via corrupting tail entity $(h, r, neg_t)$ ($neg_t$ is just a formalized symbol that represents the generated tail entity embedding at time step $t$ in this negative triple) as instance, the multimodal joint scoring of this generated negative triple is formalized as:
\begin{equation}
    S(h, r, neg_t) = (E(\textbf{x}_h, \textbf{x}_r, \hat{\textbf{x}}_{t}^{struc}) + E(\textbf{x}_h, \textbf{x}_r, \hat{\textbf{x}}_t^{vis}) +E(\textbf{x}_h, \textbf{x}_r, \hat{\textbf{x}}_t^{text})) / 3 \label{eq16}
\end{equation}
where $\hat{\textbf{x}}_{t}^{struc}$, $ \hat{\textbf{x}}_t^{vis}$ and $\hat{\textbf{x}}_t^{text}$ are the previously defined three modality-specific entity embeddings. $E(\cdot)$ represents the score function of any KGE model to evaluate the plausibility of a triple.

Furthermore, we introduce HAL which dynamically adjusts the margin based on the hardness level of negative triples generated by the DiffHEG module. This loss assigns smaller margins to harder negatives and larger margins to easier ones, enabling the model to learn effectively from challenging cases while maintaining robustness. The hardness-adaptive loss is defined as:
\begin{equation}
    \mathcal{L}_{HA} = -log \sigma(\gamma_t-E(h, r, t))- \frac{1}{|G^{-}|}\sum_{neg_t\in G^{-}}{w(neg_t) \cdot log \sigma ((S(h, r, neg_t)-\gamma_t)} \label{eq17}
\end{equation}
where $\gamma_t$ is the margin adaptive to the hardness level $HL(t)$, and \(w(neg_t)\) are the weights assigned to each negative triple. $\sigma$ represents the sigmoid function. $|G^{-}|$ is the size of the negative entity embedding set corresponding to each positive triple $(h, r, t)$. Besides, the negative triples obtained via the sampling techniques such as randomly sampling and Bernoulli sampling could be also leveraged for training KGE models with the traditional KGC loss:
\begin{equation}
   \mathcal{L}_{KGC} = -log \sigma(\gamma-E(h, r, t))- \frac{1}{|S^{-}|}\sum_{t'\in S^{-}}{log \sigma ((S(h, r, t')-\gamma)} \label{eq18}
\end{equation}
in which $\gamma$ indicates the fixed margin. $t'$ is a corrupted entity in the negative triple set $S^{-}$ obtained from previous NS techniques. The total loss for training a KGE model is a combination of \(\mathcal{L}_{KGC}\) and \(\mathcal{L}_{HA}\), weighted by a hyper-parameter $\lambda$ for a trade-off between generated and sampled negative triples:
\begin{equation}
    \mathcal{L} = \mathcal{L}_{KGC} + \lambda \cdot \mathcal{L}_{HA} \label{eq19}
\end{equation}

For a comprehensive understanding of the training procedure of our DHNS framework, the pseudo-code of training DHNS is provided in Algorithm~\ref{alg:DHNS}. 

\begin{algorithm}[H]
\caption{Training Procedure of DHNS}
\label{alg:DHNS}
\KwIn{A batch of positive triples $\mathcal{T}$ from $\mathcal{G}$ and the corresponding pre-sampled negative triples $S^{-}$ via a previous NS strategy, the pre-trained visual and textual embeddings of entities encoded from $\mathcal{M}$.}
\KwOut{The MMKGC model trained via DHNS.}

\For{each triple $(e_h, r, e_t) \in \mathcal{T}$}{
    // \textbf{Training DiffHEG module} \\
    Obtain the original modality-specific embeddings of $(e_h, r, e_t)$: $\textbf{x}_{eh}^{struc}$, $\textbf{x}_r$, $\textbf{x}_{et}^{struc}$, $\textbf{x}_{eh}^{vis}$, $\textbf{x}_{et}^{vis}$, $\textbf{x}_{eh}^{text}$, $\textbf{x}_{et}^{text}$; \\
    Calculate the noised entity embeddings at the time step $t$ as in Eq.~\ref{eq3}: $\textbf{x}_{eh,t}^{struc} = \sqrt{\bar{\beta}_t} \textbf{x}_{eh}^{struc} + \sqrt{1-\bar{\beta}_t} \epsilon_t$, $\textbf{x}_{et,t}^{struc} = \sqrt{\bar{\beta}_t} \textbf{x}_{et}^{struc} + \sqrt{1-\bar{\beta}_t} \epsilon_t$; \\
    $\textbf{x}_{eh,t}^{vis} = \sqrt{\bar{\beta}_t}\textbf{x}_{eh}^{vis} + \sqrt{1-\bar{\beta}_t} \epsilon_t$, $\textbf{x}_{et,t}^{vis} = \sqrt{\bar{\beta}_t} \textbf{x}_{et}^{vis} + \sqrt{1-\bar{\beta}_t} \epsilon_t$; \\
    $\textbf{x}_{eh,t}^{text} = \sqrt{\bar{\beta}_t}\textbf{x}_{eh}^{text} + \sqrt{1-\bar{\beta}_t} \epsilon_t$, $\textbf{x}_{et,t}^{text} = \sqrt{\bar{\beta}_t} \textbf{x}_{et}^{text} + \sqrt{1-\bar{\beta}_t} \epsilon_t$; \\
    Predict the noise at the time step $t$ following Eqs.~\ref{eq9}-\ref{eq13};\\
    Optimize parameters $\theta$ of the diffusion model by minimizing $\mathcal{L}_{diff}$; \\
    // \textbf{Generating hierarchical entity embeddings} \\
    \For{$t=T-1, \cdots, 0$}
    {
        Calculate the the modality-specific embeddings as in Eqs.~\ref{eq7}-\ref{eq8}: $neg_{eh,t}=(\hat{\textbf{x}}_{eh,t}^{struc}, \hat{\textbf{x}}_{eh,t}^{vis}, \hat{\textbf{x}}_{eh,t}^{text})$, 
        $neg_{et,t}=(\hat{\textbf{x}}_{et,t}^{struc}, \hat{\textbf{x}}_{et,t}^{vis}, \hat{\textbf{x}}_{et,t}^{text})$;\\
    }
    Generate the set of negative entity embeddings with multiple hardness levels: $G^{-}=\{(neg_{eh,t},neg_{et,t})|t=T/20, T/10, T/5, T/2\}$;\\
    // \textbf{Training a KGE model with NTAT module} \\
    Calculate the hardness-adaptive loss $\mathcal{L}_{HA}$ with $G^{-}$ as in Eq.~\ref{eq17};\\
    Calculate the KGC loss $\mathcal{L}_{KGC}$ with $S^{-}$ as in Eq.~\ref{eq18};\\
    Calculate $\mathcal{L}$ as in Eq.~\ref{eq19} and minimize it to optimize parameters of the KGE model;\\
}
\end{algorithm}

\begin{table}
\centering
\setlength{\tabcolsep}{5.5pt}
\caption{Statistics of three MMKGC benchmark datasets.}
\label{tab:datasets}
\begin{tabular}{c|c|c|c|c|c|c|c}
\toprule
Dataset & \#Entity & \#Relation & \#Image & \#Text & \#Train & \#Valid & \#Test \\
\midrule
DB15K & 12842 & 279 & 12818 & 9078 & 79222 & 9902 & 9904 \\
MKG-W & 15000 & 169 & 14463 & 14123 & 34196 & 4276 & 4274 \\
MKG-Y & 15000 & 28  & 14244 & 12305 & 21310 & 2665 & 2663 \\
\bottomrule
\end{tabular}
\end{table}

\section{Experiments}

\subsection{Experiment Settings}

\subsubsection{MMKGC Datasets.}
We conduct experiments on three MMKGC benchmark datasets: DB15K~\cite{DB15K}, MKGW and MKG-Y~\cite{MKGW}. The triples stored in DB15K are extracted from DBpedia while the structural knowledge in MKG-W (Multimodal-Wikidata) and MKG-Y (Multimodal KG-YAGO) are extracted from Wikidata~\cite{Wikidata} and YAGO~\cite{Yago}, respectively. The images in these three datasets are collected by image search engines and the textual descriptions are obtained from DBpedia. The statistics of these datasets are declared in Table~\ref{tab:datasets}.

\subsubsection{Baseline Models.}
We select three categories of baseline approaches to compare and evaluate the performance of our DHNS framework. 

(1) \textbf{Unimodal KGE models}: we select some typical KGE models learning structural features to evaluate triples, including TransE~\cite{Bordes:TransE}, TransD~\cite{TransD}, DistMult~\cite{Distmult}, ComplEx~\cite{Trouillon:ComplEx}, RotatE~\cite{RotatE}, PairRE~\cite{PairRE}, and GC-OTE~\cite{GC-OTE}. 

(2) \textbf{Multimodal KGC models}: we compare our framework with some state-of-the-art MMKGC models that could learn the multimodal features (visual features and/or textual features) together with structural features to represent a triple, including IKRL~\cite{Xie:IKRL}, TBKGC~\cite{TBKGC}, TransAE~\cite{TransAE}, MMKRL~\cite{MMKRL}, RSME~\cite{RSME}, VBKGC~\cite{VBKGC}, OTKGE~\cite{OTKGE} and AdaMF~\cite{AdaMF}. 

(3) \textbf{Negative sampling-based models}: some NS-based models are selected for evaluating on both MMKGC and NS performances, including uniform sampling (Uniform)~\cite{Bordes:TransE}, Bernoulli sampling (Bern)~\cite{Wang:TransH}, NSCaching (NSCach)~\cite{NSCaching}, KBGAN~\cite{kbgan}, SANS~\cite{SANS}, NS-KGE~\cite{NS-KGE}, MANS~\cite{MANS} and MMRNS~\cite{MMRNS}.

\subsubsection{Implementation Details.}

We implement the DHNS model with Pytorch and conduct the experiments on one NVIDIA GeForce 4090 GPU. For a fair comparison, the preprocess procedures of extracting visual and textual features are the same as AdaMF~\cite{AdaMF} and MMRNS~\cite{MMRNS}. The visual features are extracted via BEiT~\cite{beit} and the textual features are extracted using SBERT~\cite{SBERT}. Particularly, hyper-parameters of KGE models are fixed following state-of-the-art models for a fair comparison while those of our DiffHEG module are tuned. Specifically, the total timestep is selected from $\{20, 50, 70, 100\}$, learning rate is tuned in $\{2e^{-3}, 1e^{-4}, 5e^{-4}\}$. We employ two frequently-used metrics for evaluation: mean reciprocal rank of all the correct instances (MRR) and proportion of the correct instances ranked in the top $N$ (H$N$, $N=1, 3, 10$). Particularly, we employ the filter setting~\cite{Bordes:TransE} that removes the candidate triples already observed in the training set for all the results. The results of baselines in Table~\ref{tab:results_db15k_mkgw} and Table~\ref{tab:results_mkg-y} are obtained from~\cite{AdaMF} and the results in Table~\ref{tab:NS} are derived from~\cite{MMRNS}.

\subsection{Experimental Results}

\subsubsection{Main Results.}

The experimental results shown in Table~\ref{tab:results_db15k_mkgw} and Table~\ref{tab:results_mkg-y} show that our proposed framework DHNS integrated with RotatE achieves the highest or second-highest scores in terms of all the metrics across all datasets. Specifically, DHNS consistently outperforms a variety of state-of-the-art baseline models, including unimodal KGC and MMKGC models as well as NS-based models, illustrating the superior performance of our proposed framework DHNS consisting of DiffHEG and NTAT modules on MMKGC tasks.

\definecolor{lightblue}{rgb}{0.68, 0.85, 0.90}
\definecolor{lightgreen}{rgb}{0.88, 1.0, 0.88}

\begin{table}[ht]
\centering
\caption{Evaluation results (\%) of \colorbox{lightgreen}{our framework DHNS} integrated with RotatE model and some state-of-the-art baseline models on MMKGC datasets DB15K and MKG-W. The best results are marked \textbf{bold} and the second-best results are \underline{underlined}.}
\label{tab:results_db15k_mkgw}
\setlength{\tabcolsep}{3pt}
\begin{tabular}{cc|c|c|c|c|c|c|c|c}
\toprule
\multicolumn{2}{c|}{\multirow{2}{*}{Model}} & \multicolumn{4}{c|}{DB15K} & \multicolumn{4}{c}{MKG-W} \\\
& & MRR & H1 & H3 & H10 & MRR & H1 & H3 & H10 \\
\midrule
&TransE & 24.86 & 12.78 & 31.48 & 47.07 & 29.19 & 21.06 & 33.20 & 44.23 \\
&TransD & 21.52 & 8.34 & 29.93 & 44.24 & 26.84 & 19.65 & 31.48 & 42.68 \\
Unimodal & DistMult & 23.03 & 14.78 & 26.98 & 40.59 & 20.95 & 15.89 & 22.88 & 36.80 \\
KGC Models & ComplEx & 27.48 & 18.13 & 31.57 & 45.37 & 28.09 & 21.45 & 30.89 & 44.77 \\
&RotatE & 29.28 & 17.87 & 36.12 & 49.66 & 30.79 & 21.98 & 36.42 & 46.73 \\
&PairRE & 31.13 & 21.67 & 36.91 & \underline{51.98} & 33.29 & 25.00 & \underline{38.67} & 46.71 \\
&GC-OTE & 31.85 & 22.11 & 36.52 & 51.18 & 33.92 & 26.55 & 35.96 & 46.05 \\
\midrule
& IKRL & 26.82 & 14.09 & 34.93 & 49.09 & 32.36 & 26.11 & 34.75 & 44.07 \\
& TBKGC & 28.08 & 15.61 & 37.03 & 49.86 & 31.80 & 25.31 & 33.91 & 44.58 \\
& TransAE & 28.09 & 21.25 & 37.17 & 49.17 & 30.01 & 21.23 & 34.91 & 44.72 \\
MMKGC & MMKRL & 26.81 & 13.85 & 35.01 & 49.39 & 29.42 & 22.54 & 31.49 & 43.44 \\
Models & RSME & 29.76 & \textbf{24.15} & 32.12 & 49.23 & 29.23 & 23.31 & 31.09 & 40.83 \\
& VBKGC & 30.61 & 19.75 & 37.18 & 49.41 & 30.69 & 24.55 & 33.07 & 44.62 \\
& OTKGE & 23.86 & 18.45 & 25.89 & 34.23 & 34.36 & 28.85 & 36.25 & 44.88 \\
& AdaMF & \underline{32.51} & 21.31 & \underline{39.67} & 51.68 & 34.27 & 27.21 & 37.86 & 47.21 \\
\midrule
& KBGAN & 25.73 & 9.91 & 36.95 & 51.93 & 29.47 & 22.21 & 34.87 & 40.64 \\
NS-based & MANS & 28.82 & 16.87 & 38.54 & 51.51 & 32.04 & 27.48 & 37.48 & 41.62 \\
Models & MMRNS & 29.67 & 17.89 & 36.86 & 51.01 & \underline{34.47} & \underline{28.93} & 38.63 & \underline{47.48} \\
\rowcolor{lightgreen}
\cellcolor{white} & DHNS & \textbf{34.36} & \underline{23.34} & \textbf{41.56} & \textbf{53.72} & \textbf{35.68} & \textbf{28.98} & \textbf{38.68} & \textbf{48.12} \\
\bottomrule
\end{tabular}
\end{table}

\begin{table}[ht]
\centering
\caption{Evaluation results (\%) of \colorbox{lightgreen}{our framework DHNS} and some state-of-the-art baseline models on MMKGC dataset MKG-Y.}
\label{tab:results_mkg-y}
\setlength{\tabcolsep}{5pt}
\begin{tabular}{cc|c|c|c|c}
\toprule
\multicolumn{2}{c|}{\multirow{2}{*}{Model}} & \multicolumn{4}{c}{MKG-Y} \\
& & MRR & H1 & H3 & H10 \\
\midrule
& TransE & 30.73 & 23.45 & 35.18 & 43.37 \\
& TransD & 26.39 & 17.01 & 33.05 & 40.41 \\
Unimodal & DistMult & 25.04 & 19.32 & 27.80 & 39.95 \\
KGC Models & ComplEx & 28.94 & 23.11 & 31.07 & 43.48 \\
& RotatE & 29.96 & 20.70 & 36.38 & 46.49 \\
& PairRE & 32.18 & 25.24 & 37.58 & 44.98 \\
& GC-OTE & 32.95 & 26.77 & 36.44 & 44.08 \\
\midrule
& IKRL & 33.22 & 30.37 & 34.28 & 38.26 \\
& TBKGC & 33.39 & 30.47 & 33.74 & 37.92 \\
& TransAE & 28.10 & 25.31 & 29.19 & 33.03 \\
MMKGC & MMKRL & 30.94 & 27.00 & 32.53 & 36.07 \\
Models & RSME & 34.44 & 31.78 & 36.07 & 39.79 \\
& VBKGC & 34.03 & 31.76 & 35.73 & 37.72 \\
& OTKGE & 35.51 & 31.97 & 37.18 & 41.38 \\
& AdaMF & \underline{38.06} & \underline{33.49} & \underline{40.40} & \underline{45.48} \\
\midrule
& KBGAN & 29.71 & 22.81 & 34.88 & 40.21 \\
NS-based & MANS & 29.93 & 25.25 & 31.35 & 34.49 \\
Models & MMRNS & 33.32 & 30.50 & 35.37 & 45.47 \\
\rowcolor{lightgreen}
\cellcolor{white} & DHNS & \textbf{39.11} & \textbf{34.70} & \textbf{41.23} & \textbf{46.66} \\
\bottomrule
\end{tabular}
\end{table}

Besides, MMKGC models generally outperform unimodal KGC and baseline NS-based models, demonstrating the importance of incorporating supplementary multimodal information for representing KG embeddings. Notably, our framework DHNS performs better than both MMKGC and NS-based models, verifying a more effective strategy through its NS mechanism with hierarchical semantics and multiple hardness levels, together with the negative triple-adaptive training paradigm to facilitate more effective training of KGE models.


\subsubsection{Comparison of NS Strategies.}


The experimental results presented in Table~\ref{tab:NS} demonstrate that our DHNS is a robust and effective NS strategy for enhancing the performance of KGE models including TransE, DistMult and RotatE for MMKGC tasks. Specifically, DHNS consistently and significantly outperforms other NS strategies when integrated with RotatE on all the datasets and metrics. Besides, DHNS achieves the highest or near-highest MRR and H10 scores across KGE models TransE and DistMult as to all the datasets. 

More interestingly, DHNS consistently outperforms traditional and state-of-the-art NS strategies including MMRNS and also employs multimodal information across various KGE models and datasets, illustrating the superiority of directly generating hierarchical embeddings to compose diverse and high-quality negative triples rather than sampling as in baseline NS strategies. This suggests that DHNS could be regarded as a pluggable module to generate more high-quality negative triples and further guide the training of KGE models more effectively for distinguishing between positive and negative triples.

\begin{table}[htbp]
\centering
\setlength{\tabcolsep}{2pt}
\renewcommand{\arraystretch}{0.9}
\caption{Performance comparison of \colorbox{lightgreen}{our NS-based model DHNS} and various state-of-the-art NS strategies on DB15K, MKG-W, and MKG-Y.}
\begin{tabular}{c|c|ccc|ccc|ccc}
\toprule
KGE  & NS & \multicolumn{3}{c|}{DB15K} & \multicolumn{3}{c|}{MKG-W} & \multicolumn{3}{c}{MKG-Y} \\
Models & Strategies & MRR & H1 & H10 & MRR & H1 & H10 & MRR & H1 & H10 \\
\midrule
\multirow{6}{*}{TransE}
 & Uniform & 24.86 & 12.78 & 47.07 & 29.19 & 21.06 & 44.23 & 30.73 & 23.45 & 43.37 \\
 & Bern & 30.11 & 19.04 & 49.86 & 28.98 & 22.99 & 40.11 & 31.12 & 25.69 & 43.61 \\
 & KBGAN & 24.00 & 5.47 & 50.36 & 24.21 & 14.51 & 40.45 & 26.92 & 19.90 & 37.41 \\
 & NSCach & \textbf{33.03} & \textbf{23.31} & 50.29 & 28.90 & 22.24 & 40.97 & 29.51 & 24.78 & 38.02 \\
 & SANS & 26.33 & 13.87 & 48.80 & 30.22 & 23.64 & 44.61 & 27.51 & 22.31 & 42.71 \\
 & MMRNS & 26.61 & 13.40 & \underline{50.60} & \underline{33.51} & \underline{26.25} & \textbf{47.11} & \underline{34.81} & \underline{28.59} & \textbf{44.76} \\
 \rowcolor{lightgreen}
 \cellcolor{white} & DHNS & \underline{32.63} & \underline{20.95} & \textbf{52.86} & \textbf{33.83} & \textbf{26.46} & \underline{46.85} & \textbf{36.68} & \textbf{32.12} &\underline{43.62} \\
\midrule
\multirow{6}{*}{DistMult}
 & Uniform & 20.99 & 14.78 & 39.59 & 20.99 & 15.93 & \underline{38.06} & 25.04 & 19.33 & 35.95 \\
 & Bern & \underline{27.05} & \textbf{20.04} & 40.37 & 25.76 & 21.02 & 36.16 & 32.00 & \underline{30.42} & 38.11 \\
 & KBGAN & 23.36 & 16.43 & 36.30 & 19.11 & 7.69 & 37.44 & 16.99 & 9.25 & 33.33 \\
 & NSCach & 24.68 & 19.33 & 39.74 & 24.70 & 20.31 & 37.24 & 27.30 & 23.55 & 35.67 \\
 & SANS & 23.62 & 16.82 & 39.76 & 23.21 & 18.46 & 32.20 & 23.62 & 16.82 & 39.76 \\
 & MMRNS & 25.92 & 14.06 & \underline{40.42} & 25.92 & 19.73 & \textbf{38.20} & \underline{32.78} & 27.99 & \underline{41.13} \\
 \rowcolor{lightgreen}
 \cellcolor{white} & DHNS & \textbf{27.20} & \underline{19.07} & \textbf{43.01} & \underline{25.86} & \underline{20.33} & 36.53 & \textbf{37.01} & \textbf{38.72} & \textbf{42.53}\\
\midrule
\multirow{6}{*}{RotatE}
 & Uniform & 29.28 & 17.87 & 49.66 & 33.67 & 26.80 & 46.73 & 34.95 & 29.10 & 45.30 \\
 & Bern & 20.46 & 12.83 & 34.37 & 29.65 & 24.58 & 38.80 & 33.63 & 30.39 & 39.29 \\
 & NSCach & 20.32 & 12.89 & 34.05 & 32.86 & \underline{27.86} & 41.86 & 32.03 & 29.57 & 36.32 \\
 & SANS & \underline{30.51} & \underline{19.13} & 50.72 & 33.32 & 27.35 & 44.67 & 35.28 & 29.29 & 44.93 \\
 & MMRNS & 29.67 & 17.89 & \underline{51.01} & \underline{34.13} & 27.37 & \underline{47.48} & \underline{35.93} & \underline{30.53} & \underline{45.47} \\
 \rowcolor{lightgreen}
 \cellcolor{white} & DHNS & \textbf{34.36} & \textbf{23.34} & \textbf{53.72} & \textbf{35.68} & \textbf{28.98} & \textbf{48.12} & \textbf{39.11} & \textbf{34.70} & \textbf{46.66} \\
\bottomrule
\end{tabular}
\label{tab:NS}
\end{table}

\subsubsection{Ablation Study.}

We conduct an ablation study to provide insights into the effectiveness of each component of our framework DHNS integrated with RotatE on the three datasets. The ablated models can be classified from two aspects: 
\begin{itemize}
    \item To verify each contribution of our developed NS strategy, the ablated models are constructed by removing the whole DiffHEG module (w/o DiffHEG) or two sub-modules namely multimodal conditioning (w/o MMC) and multiple hardness-level denoising (w/o MHLD).
    \item Specific to our training strategy, the ablated models are designed by removing the whole NTAT module (w/o NTAT) or the key sub-module namely the hardness-adaptive loss (w/o HAL) while maintaining the hardness-aware weights of generated negative triples. 
\end{itemize}

\begin{table}[htbp]
\centering
\setlength{\tabcolsep}{2.5pt}
\caption{Ablation study of our framework on DB15K, MKG-W, and MKG-Y datasets. The best results are \textbf{bold} and the lowest values are labeled with the superscript$^*$.}
\begin{tabular}{l|ccc|ccc|ccc}
\toprule
\multirow{2}{*}{Ablated Models} & \multicolumn{3}{c|}{DB15K} & \multicolumn{3}{c|}{MKG-W} & \multicolumn{3}{c}{MKG-Y} \\
& MRR & H1 & H3 & MRR & H1 & H3 & MRR & H1 & H3 \\
\midrule
DHNS & \textbf{34.36} & \textbf{23.34} & \textbf{41.56} & \textbf{35.68} & \textbf{28.98} & \textbf{38.68} & \textbf{39.11} & \textbf{34.70} & \textbf{41.23} \\
\midrule
w/o DiffHEG & 30.91$^*$ & 20.96$^*$ & 35.85$^*$ & 29.02$^*$ & 23.10$^*$ & 31.77$^*$ & 35.50$^*$ & 30.32$^*$ & 36.86$^*$ \\
\ \ \ \ \ w/o MMC & 32.82 & 22.67 & 37.79 & 31.49 & 26.01 & 33.76 & 35.75 & 33.06 & 39.02 \\
\ \ \ \ \ w/o MHLD & 31.67 & 21.71 & 37.80 & 30.88 & 25.55 & 32.88 & 36.26 & 32.69 & 39.86 \\
\midrule
w/o NTAT & 32.59 & 21.29 & 37.59 & 30.78 & 25.56 & 32.92 & 36.67 & 33.68 & 39.28 \\
\ \ \ \ \ w/o HAL & 33.78 & 22.05 & 38.31 & 31.22 & 25.80 & 33.26 & 36.77 & 34.54 & 39.67 \\
\bottomrule
\end{tabular}
\label{tab:ablation}
\end{table}

From the results shown in Table~\ref{tab:ablation}, we could observe that the performance of each ablated model drops compared with the whole framework DHNS across all the datasets, illustrating the effectiveness of each contribution of our framework on MMKGC tasks. Particularly, removing DiffHEG leads to the most significant drops in performance across all datasets. For instance, the performance of the ablated model w/o DiffHEG decreases \textbf{11.2}\%/\textbf{22.9}\%/\textbf{10.2}\% as to MRR on DB15K/MKG-W/MKG-Y. This indicates that our proposed DiffHEG module for generating negative triples plays a crucial role in improving the model's ability on MMKGC tasks. Besides, we could demonstrate that both multimodal conditioning and multiple hardness level denoising mechanisms are effective for generating diverse and high-quality negative triples from the perspectives of multimodal semantics and multiple hardness levels.

More interestingly, compared with eliminating the NS module DiffHEG, removing the training strategy NTAT exhibits a more slight decrease in performance on MKG-Y while similar significant drops on DB15K and MKG-W. In specific, the performance of the ablated model w/o NTAT drops \textbf{5.4}\%/\textbf{15.9}\%/ \textbf{6.7}\%  as to MRR on DB15K/MKG-W/MKG-Y. Furthermore, the ablated model w/o HAL shows a similar performance to w/o NTAT, indicating that HAL plays a key role in the training strategy to adaptively select the margin to improve the capability of our model in discriminating positive and negative with various hardness levels. In summary, both our proposed NS and training strategies contribute to the performance improvement of the MMKGC model.

\section{Conclusion}

In this paper, we propose a novel Diffusion Model-based Hierarchical Negative Sampling framework DHNS for MMKGC tasks. To address the unique challenge of uncontrollable negative sampling, especially in the context of MMKGs, we are the first to develop a Diffusion-based Hierarchical Embedding Generation (DiffHEG) module to directly generate hierarchical embeddings rather than entity sampling to compose negative triples, with multimodal semantics and varying hardness levels determined by the diffusion time steps. Then, to handle the issue of the traditional one-margin-fits-all training scheme, a Negative Triple-Adaptive Training (NTAT) strategy is designed to learn the multimodal joint scoring of the generated negative triples, and further train KGE models with a Hardness-Adaptive Loss (HAL) to improve the discrimination capability concerning the diversity among negative triples. The extensive results on three MMKGC datasets illustrate the effectiveness and superiority of our DHNS framework compared with several state-of-the-art unimodal and multimodal KGC models as well as some typical NS techniques.

%
%
%
\bibliographystyle{splncs04}
\bibliography{ref}
%
\end{document}